%% file: main.tex
\documentclass{article}
\usepackage{styles/spconf,amsmath,graphicx}
\usepackage{hyperref}
\title{IMPROVED HARD EXAMPLE MINING APPROACH FOR SINGLE SHOT OBJECT DETECTORS}
\name{Aybora K\"{o}ksal$^{\star}$ \quad \"{O}nder Tuzcuo\u{g}lu$^{\star}$ \quad Kutalm{\i}\c{s} G\"{o}kalp \.{I}nce$^{\star}$ \quad Yolda\c{s} Ataseven$^{\dagger}$ \quad A. Ayd{\i}n Alatan$^{\star}$\thanks{This study is funded by ASELSAN Inc. The codes are available at \href{https://github.com/aybora/yolov5Loss}{github.com/aybora/yolov5Loss}}}
\address{$^{\star}$ Center for Image Analysis (OGAM), Department of Electrical and Electronics Engineering\\ Middle East Technical University, Ankara, Turkey\\
$^{\dagger}$ ASELSAN Inc., Ankara, Turkey}
\begin{document}
%
\maketitle
\begin{abstract}
Hard example mining methods generally improve the performance of the object detectors, which suffer from imbalanced training sets. In this work, two existing hard example mining approaches (LRM and focal loss, FL) are adapted and combined in a state-of-the-art real-time object detector, YOLOv5. The effectiveness of the proposed approach for improving the performance on hard examples is extensively evaluated. The proposed method increases mAP by 3\% compared to using the original loss function and around 1-2\% compared to using the hard-mining methods (LRM or FL) individually on 2021 Anti-UAV Challenge Dataset.
\end{abstract}

\input{chapters/introduction}
\input{chapters/method}

\input{chapters/experiments}
\input{chapters/conclusion}

\bibliographystyle{styles/IEEEbib}
\bibliography{refs}

\end{document}

%% file: chapters/introduction.tex
\vspace{-1em}
\section{Introduction}
\label{sec:intro}

Object detection performance has rapidly increased during the last decade by the utilization of Convolutional Neural Networks (CNN) for feature extraction. Even though most of the object detectors \cite{Girshick1,Girshick2,yolo,yolov5github} work well on common datasets, such as MS COCO \cite{coco}, they usually suffer from two main problems: the imbalance between the number of background-foreground data and infrequent observation of trained foreground object representations in the test set, i.e. the \textit{tail problem}. 

In order to cope with the imbalance problem, some example mining methods are proposed for the two-stage object detectors \cite{Girshick1,Girshick2,Ren,dai2016r}. All of these methods are specific to the two-stage detectors, since they are based on the outputs of the RoI Pooling stage. Therefore, these methods are not applicable to the single shot object detectors. Unfortunately, the pioneering examples of single shot object detectors \cite{yolo,ssd,redmon2016yolo9000,yolov3}, do not have a solution for this problem. Besides the imbalance problem, some appearances of a class might be rare in the dataset of interest. Such rare occurrences which lie at the tails of the appearance distribution are dominated by the rest of the dataset, and therefore, they are hard to learn.   

\begin{figure*}[ht]
\centering
   \includegraphics[width=0.8\textwidth]{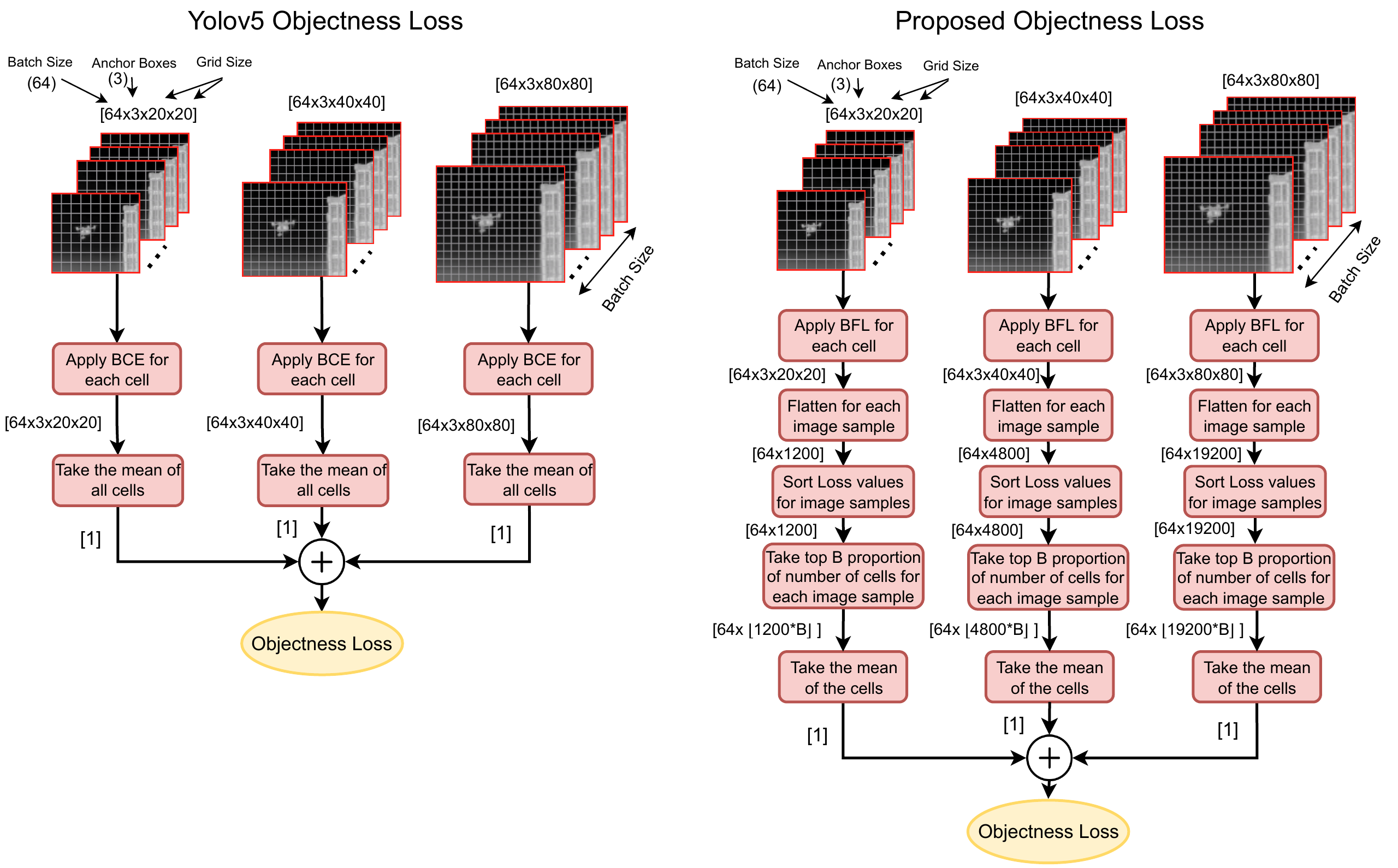}
    \vspace{-1em}
   \caption{YOLOv5 objectness loss (left) vs. proposed combined loss (right). For the proposed loss, firstly Balanced Focal Loss is applied for each cell instead of BCE, then for each feature map, detections are sorted with respect to their loss values. Finally, top $B$ detection which has highest loss values are selected for loss calculation and backpropagation.}
\label{fig:comb}
\end{figure*}

Focusing on the hard examples of an imbalanced dataset is tried to be handled by \textit{bootstrapping} by Sung \cite{sung1996learning}. The main idea of this approach is incrementally increasing the weight of the examples that trigger false alarms; this study is one of the prominent solutions for iterative learning. Later, bootstrapping ideas are also used in SVM, with the introduction of Latent SVM paradigm \cite{felzenszwalb2009object}. Finally, bootstrapping methods also became popular in object detection the object detection research by the utilization of SVM \cite{Girshick1, he2015spatial, uijlings2013selective}.

In order to mine a hard example via RoIs properly, Online Hard Example Mining (OHEM) method is introduced \cite{shrivastava2016training}. The idea suggests considering only the most beneficial RoIs for the backpropagation. RoIs which give the highest loss values are assumed to be the hardest examples, and therefore, the most beneficial ones. Hence, the aforementioned method selects $B/N$ worst loss cases for training and discards the remaining during training. Although this novel approach is one of the most promising approaches in hard example mining, it is only applicable to two-stage networks, since it requires RoIs to work on.


Lin et. al. \cite{retinanet} introduced an inherent hard example mining method for a single shot object detector without sacrificing its real time performance. They introduced focal loss to use hard examples more effectively. The loss function is designed to make the detections with higher loss values more important in back propagation than the others by performing gamma correction with a $\gamma$ factor larger than 1. After its efficiency was observed, focal loss was also used in other one stage object detectors such as EfficientDet \cite{tan2020efficientdet}. The idea was also applied to YOLOv3 on MS COCO dataset \cite{coco}, but it did not increase the baseline performance \cite{yolov3}. Since the other state-of-the-art object detectors are working well with focal loss, it might be worth trying to modify focal loss in YOLO to make it work properly.

Based on the idea of OHEM, Yu et. al. introduce Loss Rank Mining (LRM) \cite{yu2018loss}. The method is applicable for single shot detectors, and it makes the object detector to focus on hard examples by filtering-out some easy examples on the feature map just before the detection stage. During training, as the first step, the input goes through the model backbone to get the feature map. Then, for each detection, the loss value is calculated. After the non-maximum suppresson (NMS) stages, the loss values of these detections are sorted in descending order and the first $K$ detection results are selected and filtered. The rest of the detection values are not used during the training process. This idea might be applicable and beneficial to the current object detectors, if it can be implemented into their structure. 

In the recent studies, several methods for hard example mining are also proposed. Jin et. al. \cite{jin2018unsupervised} introduced an unsupervised hard example mining method for video sequences. Their approach suggests a template matching solution between consecutive frames. If the matched templates are not temporarily consistent, they are flagged as hard examples and the training continues iteratively. Wang et. al. \cite{wang2017fast} propose an adversarial network structure in order to create artificially occluded hard examples for the imbalance problem. 
Although both of these approaches are worth to mention, unfortunately, they are not automatic processes and they frequently need human intervention.

The proposed study combines two different hard example mining approaches and applies the resulting method on YOLOv5, which is one of the best-performing single shot object detectors. For that purpose, the focal loss is adapted to YOLOv5 and LRM (which is originally designed to work with a single feature map) is modified to work with multiple feature maps. Next, these two methods are combined to obtain a single loss function, as shown in Figure \ref{fig:comb}. Quantitative experiments are conducted to verify that the methods are mining the hard examples without manipulating the number of hard examples.

The proposed method differs from the previous work in the following points: a) Our modified focal loss approach  increases YOLOv5 detection performance, which is not the case for the original focal loss implementation of YOLOv5; b) Our LRM structure filters the detections in each feature map separately, which is not achieved in the original LRM; c) The proposed approach combines the focal loss and LRM approaches into one novel loss function; d) Our performance evaluation method allows us to check whether the suggested approach increases the performance on hard examples, without defining them explicitly.

%% file: chapters/method.tex
\section{Proposed Method}
\label{sec:method}

The proposed approach has a combined hard example mining structure which uses Balanced Focal Loss and Loss Rank Mining. Both of these methods are proposed in a way these can be used individually or combined. 

\subsection{Balanced Focal Loss}

The popular cross entropy loss function is shown in (\ref{eq:ce}), whereas original Focal loss function, proposed by Lin et. al. \cite{retinanet}, is given in (\ref{eq:fl}). 

\begin{equation}
     CE(p) = -\log(p) 
     \label{eq:ce}
\end{equation}

\vspace{-.5cm}

\begin{equation}
    FL(p) = -\alpha(1-p)^\gamma \log(p)
    \label{eq:fl}
\end{equation}

In (\ref{eq:fl}), $\gamma$ is the focus parameter, while $\alpha$ denotes the correction parameter. The original YOLOv5 implementation already has a flag for activation of focal loss. However, it generally decreases the performance of YOLOv5, since $\gamma$ factor makes the value of the objectness loss negligibly small so that it becomes insignificant with respect to the box regression loss. Therefore, it should be scaled appropriately in order to make these two loss values comparable. In our work, focal loss is weighted by an additional balancing parameter, $\xi$. It should be noted that $\xi$ is weight of the objectness loss in the overall loss function which is aimed to be higher than $1$. 

\subsection{Loss Rank Mining}

In its original paper \cite{yu2018loss}, this method is used with YOLOv2 \cite{redmon2016yolo9000}, which has one feature map for object detection. Since YOLOv5 uses three feature maps for small, medium and large objects, the original method is also modified in such a way that it works with all three feature maps. 

In the original LRM structure, first $K$ detection results with the highest amount of loss are selected. In our method, first $B$ (rank factor) detections are selected for each feature map.  The comparison between the proposed combined objectness loss structure and the original YOLOv5 loss is illustrated in Figure \ref{fig:comb} and the method can be summarized as follows \textit{for each feature map}:

\begin{enumerate}
    \item Through each mini-batch, Balanced Focal Loss is applied to cells for obtaining loss values of the detections.  
    
    \item By flattening the three-dimensional cell structure, loss values of each image sample are concatenated into different vectors separately.
    
    \item Loss values of each image are sorted by a value.
    
    \item From the sorted loss vectors, the top $B$ proportion of the number of cells for each image sample is selected. 
    
    \item The mean of each selected loss is taken individually.
    
    \item These averages are summed to form the objectness loss.
\end{enumerate}

%% file: chapters/experiments.tex
\section{Experiments}
\label{sec:experiments}

Throughout the experiments, YOLOv5s \cite{yolov5github} is used as the baseline detector. 2021 Anti-UAV Challenge Dataset\footnote{\url{https://anti-uav.github.io/dataset/}} is used during the experiments. In order to speed-up the training phase, the training set is generated with 1/20 of the original frame rate. Removing adjacent video frames is known not to cause a significant drop in the detection performance \cite{Koksal_2020_CVPR_Workshops}. 3818 frames are selected for the training set, whereas 2313 for the validation, and 1517 is selected for the test set. 

As an ablation study, the methods are compared \textit{head-to-head} according to their detection performance (True Positive (TP), True Negative (TN), False Positive (FP) and False Negative (FN)) considering objectness score (confidence) and Intersection-over-Union (IoU) metric. More specifically, for any detection, if confidence $>$ 0.5 and IoU $>$ 0.5, that detection is accepted as a TP. If confidence $>$ 0.5 but IoU $<$ 0.5, that is a FP. If there is a ground truth detection whose confidence $<$ 0.5, then ground truth object becomes a FN. If confidence $<$ 0.5 and there is no ground truth, that is a TN.

\begin{figure}[ht]
\centering
   \includegraphics[width=0.9\linewidth]{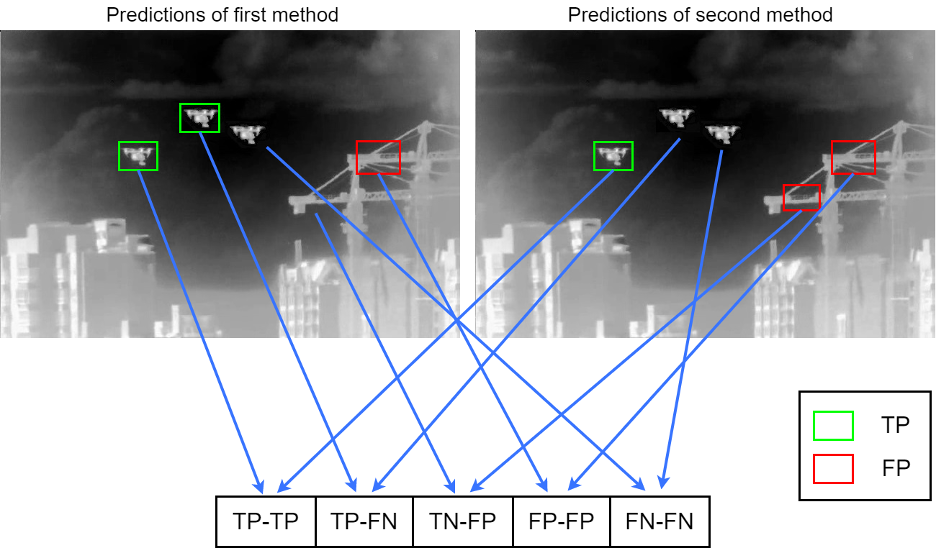}
   \caption{A prediction confusion matrix pairing method on one frame for the comparison of two different training model. Image taken from 2021 Anti-UAV dataset$^1$.}
\label{fig:pairs}
\end{figure}

As we do not have a predefined hard example set, we define hard examples as the distinguishing failures of alternative methods. The detections and misses which fall at the same cell are classified as TP-TP, TP-FN, TN-FP, FP-FP, FN-FN, as it can be observed in Figure \ref{fig:pairs}. Since all of these methods are already the state-of-the-art, the frames for which both of these algorithms fail are counted as hard examples. Therefore, the aim of these experiments is to check TP-FN and TN-FP pairs. The reason is, if method A has correct outputs (TPs and TNs) for some FNs and FPs of method B, and method A has less incorrect outputs for TPs and TNs of method B, then it is reasonable to assume that method A is better for hard examples. This is an unsupervised performance evaluation approach, since the number of hard examples is unknown.

\textbf{Default loss vs Focal Loss:} Original YOLOv5 loss function and Original Focal Loss are compared with $\gamma$: 1.5 and $\alpha$: 0.25. According to the results which are tabulated in Table \ref{tab:dvsfl}, the baseline focal loss function degrades the algorithm by mistakenly converting 8.06\% of the test set which was evaluated as TP into FN. Although it also converts some FNs into TPs and FP into TNs, the overall performance of hard examples is decreased by 6\% on test set by implementing original focal loss into YOLOv5. Therefore, the baseline focal loss is not used for the rest of the experiments.

\begin{table}[ht]
Pairwise comparison of method pairs in terms of Number (\#) and Percentage (\%) of Frames in the Test Set
\begin{minipage}[b]{.48\linewidth}
\caption{M1: Default, M2: Focal Loss} 
\centering
\label{tab:dvsfl}
\medskip
\begin{tabular}{|c|c|c|c|}
\hline
\textbf{M1} & \textbf{M2} & \textbf{\# Fr.} & \textbf{Fr. \%} \\
\hline 
FN & TP & 9 & 0.59 \\ 
FP & TN & 17 & 1.12  \\
\hline
TP & FN & 122 & 8.06 \\ 
TN & FP & 7 & 0.46 \\
\hline
FP & FP & 4 & 0.26 \\  
FN & FN & 215 & 14.20 \\
TP & TP & 1140 & 75.30  \\ 
\hline
\end{tabular}
\end{minipage}
\hfill
\begin{minipage}[b]{.48\linewidth}
\caption{M1: Default, M2: Bal. Focal Loss ($\xi: 30$)} 
\centering
\label{tab:dvssfl}
\medskip
\begin{tabular}{|c|c|c|c|}
\hline
\textbf{M1} & \textbf{M2} & \textbf{\# Fr.} & \textbf{Fr. \%} \\
\hline
FN & TP & 63 & 4.14 \\
FP & TN & 14 & 0.92 \\
\hline 
TP & FN & 12 & 0.79 \\
TN & FP & 16 & 1.05 \\
\hline
FP & FP & 7 & 0.46 \\ 
FN & FN & 161 & 10.57 \\
TP & TP & 1250 & 82.07 \\
\hline
\end{tabular}
\end{minipage}

\begin{minipage}[b]{.48\linewidth}
\caption{M1: Default, M2: LRM ($B: 0.35$)} 
\centering
\label{tab:dvslrm}
\medskip
\begin{tabular}{|c|c|c|c|}
\hline
\textbf{M1} & \textbf{M2} & \textbf{\# Fr.} & \textbf{Fr. \%} \\
\hline
FN & TP & 75 & 4.92 \\ 
FP & TN & 14 & 0.92 \\ 
\hline
TP & FN & 5 & 0.33 \\ 
TN & FP & 16 & 1.05 \\ 
\hline
FP & FP & 7 & 0.46 \\ 
FN & FN & 149 & 9.78 \\ 
TP & TP & 1257 & 82.53 \\ 
\hline
\end{tabular}
\end{minipage}
\hfill
\begin{minipage}[b]{.48\linewidth}
\caption{M1: Default, M2: Combined ($\xi: 30$, $B: 0.35$)} 
\centering
\label{tab:dvscomb}
\medskip
\begin{tabular}{|c|c|c|c|}
\hline
\textbf{M1} & \textbf{M2} & \textbf{\# Fr.} & \textbf{Fr. \%} \\
\hline
FN & TP & 83 & 5.47 \\ 
FP & TN & 16 & 1.06 \\ 
\hline
TP & FN & 9 & 0.59 \\ 
TN & FP & 9 & 0.59 \\ 
\hline
FP & FP & 5 & 0.33 \\ 
FN & FN & 141 & 9.30 \\ 
TP & TP & 1253 & 82.65 \\
\hline
\end{tabular}
\end{minipage}

\begin{minipage}[b]{.48\linewidth}
\caption{M1: LRM ($B:0.35$, M2: Combined ($\xi: 30$, $B: 0.35$)} 
\centering
\label{tab:lrmvscomb}
\medskip
\begin{tabular}{|c|c|c|c|}
\hline
\textbf{M1} & \textbf{M2} & \textbf{\# Fr.} & \textbf{Fr. \%} \\
\hline
FN & TP & 23 & 1.52 \\ 
FP & TN & 17 & 1.12 \\ 
\hline
TP & FN & 19 & 1.25 \\ 
TN & FP & 8 & 0.53 \\ 
\hline
FP & FP & 6 & 0.40 \\ 
FN & FN & 131 & 8.64 \\ 
TP & TP & 1313 & 86.55 \\ 
\hline
\end{tabular}
\end{minipage}
\hfill
\begin{minipage}[b]{.48\linewidth}
\caption{M1: Bal. Focal Loss ($\xi: 30$, M2: Combined ($\xi: 30$, $B: 0.35$)} 
\centering
\label{tab:sflvscomb}
\medskip
\begin{tabular}{|c|c|c|c|}
\hline
\textbf{M1} & \textbf{M2} & \textbf{\# Fr.} & \textbf{Fr. \%} \\
\hline
FN & TP & 34 & 2.24 \\
FP & TN & 15 & 0.99 \\
\hline
TP & FN & 11 & 0.73 \\
TN & FP & 6 & 0.40 \\
\hline
FP & FP & 8 & 0.53 \\
FN & FN & 139 & 9.17 \\
TP & TP & 1302 & 85.94 \\
\hline
\end{tabular}
\end{minipage}

\vspace{-1em}
\end{table}

\textbf{Default loss vs. Balanced Focal Loss:} Original YOLOv5 loss function and the proposed Balanced Focal Loss are compared by using $\gamma$: 1.5, $\alpha$: 0.25 and $\xi$: 30. According to the results which are shown in Table \ref{tab:dvssfl}, Balanced Focal Loss successfully converts 4.14\% of the test set to TP which was FN in default loss function, while lost its 0.79\% of TP to FN. Therefore, the performance of hard examples is improved by more than 3\% with the usage of Balanced Focal Loss.

\textbf{Default loss vs. LRM:} Original YOLOv5 loss function and LRM are compared with $B$: 0.35. According to the results in Table \ref{tab:dvslrm}, 4.92\% of the test set evaluated as FN are converted to TP, while 0.33\% of them are lost. In the overall evaluation, the performance of the hard examples is increased by around 4.50\% with LRM.

\textbf{Default loss vs. Combined:} Original YOLOv5 loss is compared with our Combined loss approach with $\gamma$: 1.5, $\alpha$: 0.25 and $\xi$: 30 and $B$: 0.35. According to the results in Table \ref{tab:dvscomb}, 5.47\% of the test set evaluated as FN are converted to TP, while 0.59\% of them are lost. Therefore, the performance of the hard examples are increased by around 5\% with LRM and Balanced Focal Loss combined.

\textbf{LRM vs Combined:} Proposed LRM is compared with our Combined loss approach with $\gamma$: 1.5, $\alpha$: 0.25 and $\xi$: 30 and $B$: 0.35.  According to the results which are presented in Table \ref{tab:lrmvscomb}, 1.52\% of the test set evaluated FN are converted to TP, while 1.25\% of them are lost. Moreover, 1.12\% of the set which was FP are converted to TN, while 0.53\% are lost. Therefore, using the balanced focal loss in addition to LRM increases the overall hard example performance by 1\%. 

\textbf{Balanced Focal Loss vs Combined:} Proposed Balanced Focal Loss is compared with our Combined loss approach with $\gamma$: 1.5, $\alpha$: 0.25 and $\xi$: 30 and $B$: 0.35. According to the results in Table \ref{tab:sflvscomb}, 2.24\% of the test set which was FN are converted to TP, while 0.73\% of them are lost. Moreover, 0.99\% of the set evaluated as FP are converted to TN, while 0.40\% are lost. Overall, using LRM in addition to Balanced Focal Loss increases the hard example performance by 2\%.

After the head-to-head comparison of algorithms for hard example performance, it is time to check their overall performance by using standard object detection metrics. For that purpose, precision, recall and mAP@.5 metrics are used. 

The results are tabulated in Table \ref{tab:map}. For these experiments, the parameters are kept constant with following values: $\alpha=0.25$,  $\gamma=1.5$, $\xi=30$, $B=0.35$. Table \ref{tab:map} indicates that the prior experiments are coherent with the experiments on object detection metrics, our LRM and Balanced Focal Loss combined approach outperforms all the other loss selections in terms of Precision, Recall and mAP@.5 metrics. 

\vspace{-1em}

\begin{table}[ht]
\caption{Performance evaluation of the baseline and proposed methods in terms of Precision (\%), Recall (\%), mAP$_{0.5}$ (\%) and mAP$_{0.5:0.95}$ (\%).} 
\centering
\label{tab:map}
\medskip
\begin{tabular}{|l|c|c|c|c|}
\hline
{\bf Method} & Prec. & Rec. & mAP$_{0.5}$ & mAP$_{0.5:0.95}$ \\
\hline
Default & 98.0 & 85.4 & 90.4 & 53.3 \\
Focal Loss & 93.6 & 84.3 & 90.4 & 53.3 \\
Bal. Focal Loss & 98.2 & 89.6 & 92.6 & 55.9 \\
LRM & 98.0 & 90.0 & 93.2 & \textbf{57.3} \\
Combined & \textbf{98.3} & \textbf{91.0} & \textbf{93.5} & 56.1 \\

\hline
\end{tabular}
\end{table}

%% file: chapters/conclusion.tex
\vspace{-1em}

\section{Conclusion}
\label{sec:conc}

Two hard example mining methods are modified and adapted on a state-of-the-art object detector, YOLOv5. The experiments clearly indicated that although the original focal loss degrades the precision of YOLOv5, the proposed Balanced Focal Loss corrects such inaccuracies and improves the overall performance. Similarly, LRM structure is modified to integrate with YOLOv5 architecture and the experiments demonstrate a meaningful increase in mAP scores. Finally, Balanced Focal Loss and LRM methods are combined and the final object detection performance is calculated as 93.5\% mAP, improving the baseline performance by 3.1\%. 